\documentclass[letterpaper, 10 pt, conference]{ieeeconf}  

\IEEEoverridecommandlockouts                              

\usepackage{tabularx}
\usepackage{multirow}
\usepackage{booktabs}
\usepackage{subcaption}
\usepackage{graphicx}
\usepackage{caption}
\usepackage{array}
\usepackage{cite}
\usepackage{hyperref}
\graphicspath{{images/}}
\usepackage{xcolor}

\title{\LARGE \bf MARBLER: An Open Platform for Standardized Evaluation of Multi-Robot Reinforcement Learning Algorithms}

\author{
  Reza J. Torbati\thanks{$^\dag$Denotes equal contribution; This work was supported in part by the Army Research Lab under Grants W911NF-17-2-0181 (DCIST CRA) and W911NF-20-2-0036},
  Shubham Lohiya$^\dag$,
  Shivika Singh$^\dag$,
  Meher S. Nigam,
  Harish Ravichandar
}

\begin{document}
\maketitle

\begin{abstract} 
Multi-Agent Reinforcement Learning (MARL) has enjoyed significant recent progress thanks, in part, to the integration of deep learning techniques for modeling interactions in complex environments. This is naturally starting to benefit multi-robot systems (MRS) in the form of multi-robot RL (MRRL). However, existing infrastructure to train and evaluate policies predominantly focus on the challenges of coordinating virtual agents, and ignore characteristics important to robotic systems. Few platforms support realistic robot dynamics, and fewer still can evaluate Sim2Real performance of learned behavior. To address these issues, we contribute MARBLER: \textbf{M}ulti-\textbf{A}gent \textbf{R}L \textbf{B}enchmark and \textbf{L}earning \textbf{E}nvironment for the \textbf{R}obotarium. MARBLER offers a robust and comprehensive evaluation platform for MRRL by marrying Georgia Tech's Robotarium (which enables rapid deployment on physical MRS) and OpenAI's Gym interface (which facilitates standardized use of modern learning algorithms). MARBLER offers a highly controllable environment with realistic dynamics, including barrier certificate-based obstacle avoidance. It allows anyone across the world to train and deploy MRRL algorithms on a physical testbed with reproducibility. Further, we introduce five novel scenarios inspired by common challenges in MRS and provide support for new custom scenarios. Finally, we use MARBLER to evaluate popular MARL algorithms and provide insights into their suitability for MRRL. In summary, MARBLER can be a valuable tool to the MRS research community by facilitating comprehensive and standardized evaluation of learning algorithms on realistic simulations and physical hardware. Links to our open-source framework and videos of real-world experiments can be found at \url{https://shubhlohiya.github.io/MARBLER/}.

\end{abstract}

\section{Introduction}
With increasing demands for robotics to solve complex real-world challenges, coordination of multiple robots is becoming paramount. However, the complexity of exact solutions to important problems (e.g., coverage control~\cite{h2gnn}, path-planning~\cite{li2020graph}, and task allocation~\cite{neville2021interleaved}) grows exponentially as the number of robots increase~\cite{complexity}. Consequently, Multi-Robot Reinforcement Learning (MRRL)~\cite{mataric1997reinforcement} is emerging as a promising alternative paradigm to address these challenges.

MRRL has proven useful for delivery robots~\cite{delivery}, coordinated robotic exploration~\cite{h2gnn}, multi-robot communication~\cite{hetnet,communication}, multi-robot path planning~\cite{mapf}, multi-robot target localization~\cite{alagha2022target} and more~\cite{marlsurvey}. However, despite being developed for robotics, learning algorithms are rarely evaluated in the real-world, with a few notable exceptions~\cite{hetgppo,yu2023asynchronous,lin2019end,han2020cooperative}. However, even the exceptions were tested on smaller teams (2, 2, 3, and 4 robots, respectively) and on ad-hoc platforms, rendering reproducibility time-consuming and difficult.

In contrast, Multi-\textit{Agent} Reinforcement Learning (MARL) algorithms tend to be evaluated in a systematic way in many standardized simulated environments, such as the Multi-Agent Particle Environment (MPE)~\cite{MPE} and the StarCraft Multi-Agent Challenge (SMAC)~\cite{smac}. While it might be possible to use existing MARL environments to evaluate algorithms developed for MRS, they lack realistic robot dynamics which likely leads to a large Sim2Real gap.

\begin{figure}
  \centering
  \includegraphics[width=\columnwidth]{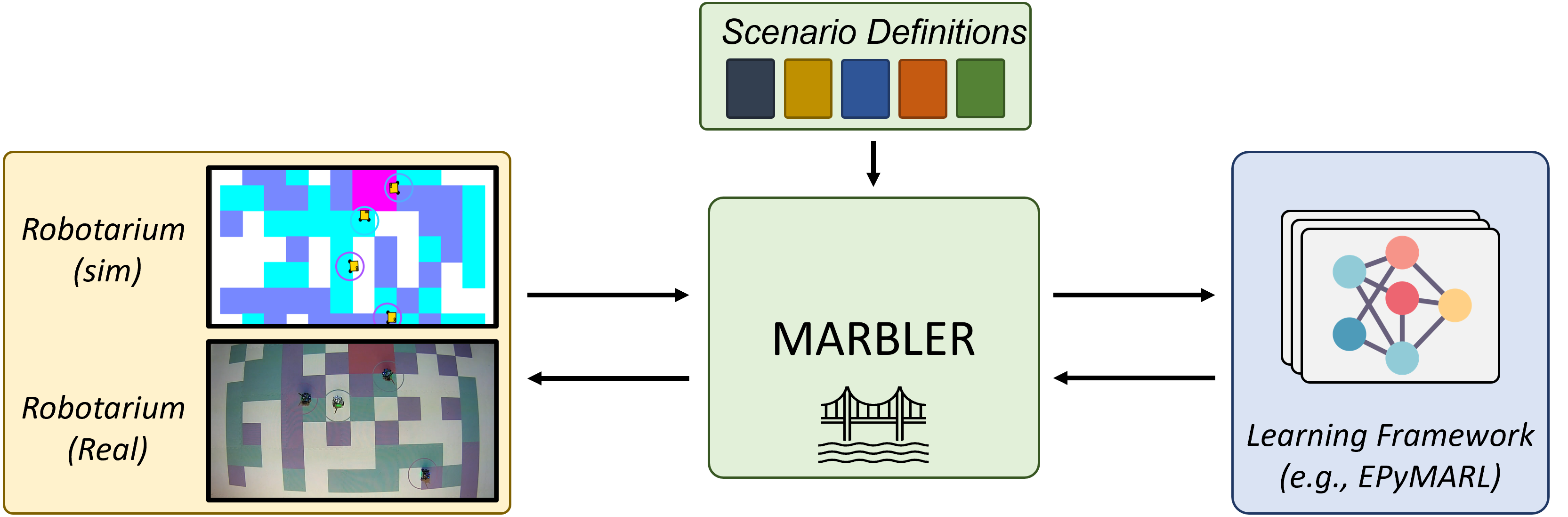}
  \caption{\small{
  MARBLER enables users to train coordination policies with an explicit emphasis on multi-robot teams by serving as a bridge between the state-of-the-art in MARL algorithms (e.g., EPyMARL~\cite{papoudakis2021benchmarking}) with that in multi-robot testbed (Robotarium~\cite{robotarium}). 
  }}
  \label{fig:method}
\end{figure}

In this work, we develop a holistic, open-source platform that can enable seamless training of MRRL policies and their evaluation on physical robots. Specifically, we contribute \textbf{M}ulti-\textbf{A}gent \textbf{R}L \textbf{B}enchmark and \textbf{L}earning \textbf{E}nvironment for the \textbf{R}obotarium (MARBLER). MARBLER is a bridge between the MARL community and the physical robots in the Robotarium~\cite{robotarium} that makes it easy to evaluate MRRL algorithms and design novel scenarios. The Robotarium is a remotely-accessible, publicly-available, and free-to-use testbed for MRS that allows for up to 20 robots at once in a highly-customizable environment.  
As such, MARBLER enables machine learning researchers to develop and test algorithms for physical robots, and control theorists to experiment with state-of-the-art (SOTA) learning algorithms. 

\noindent Our MARBLER platform has the following key benefits:
\begin{enumerate}
    \item The simulated robots in MARBLER exhibit dynamics similar to that of physical robots as it is built on top of the Robotarium's simulator. Further, MARBLER includes support for barrier certificates to prevent collisions, forcing algorithms to learn in realistic settings.
    \item MARBLER inherits the open-access benefits of the Robotarium, enabling anyone across the world to train coordination algorithms and systematically deploy on a physical multi-robot testbed with reproducibility.
    \item MARBLER is compatible with any learning algorithm that can be used with the OpenAI Gym interface.
    \item MARBLER currently has 5 novel scenarios inspired by common and challenging problems in MRS.
    \item MARBLER is open-source and allows users to easily add new scenarios or modify existing ones.
\end{enumerate}

By creating an interface between MARL algorithms and the Robotarium, MARBLER is the first publicly-available environment that can evaluate Sim2Real capability in MRRL. Further, MARBLER can serve as a benchmark to evaluate learning algorithms in simulation with real-world constraints and readily deploy them on physical robots. 

In addition, we conducted detailed evaluations of existing MARL algorithms by leveraging Extended PyMARL (EPyMARL)~\cite{papoudakis2021benchmarking} within MARBLER. Our experiments reveal insights into how different characteristics of existing algorithms (e.g., policy gradient vs. valued-based, parameter sharing, etc.) impact performance in both simulated and physical multi-robot systems.

\section{Related Work}

\subsection{MARL and MRRL Platforms} 

\begin{table*}[h]
\renewcommand{\arraystretch}{1.3} 
\setlength{\tabcolsep}{3.2pt} 
\begin{tabular}{|>{\centering\arraybackslash}m{2.5cm}|>{\centering\arraybackslash}m{1.8cm}|>{\centering\arraybackslash}m{2.3cm}|>{\centering\arraybackslash}m{1.8cm}|>{\centering\arraybackslash}m{2cm}|>{\centering\arraybackslash}m{1.8cm}|>{\centering\arraybackslash}m{1.8cm}|>{\centering\arraybackslash}m{1.8cm}|}
\hline
\textbf{Platform} & \textbf{Robot-based Dynamics} & \textbf{Collision Avoidance} & \textbf{OpenAI Gym Compatibility} & \textbf{Max \#Agents/Robots} & \textbf{Sim2Real Capabilities} & \textbf{Public Testbed Available} & \textbf{Custom Scenarios} \\
\hline
MPE~\cite{MPE}   & No & Optional (elastic) & Yes & No limit & No  & N/A & Yes \\ \hline
VMAS~\cite{vmas}   & No & Optional (elastic) & Yes & No limit & No  & N/A & Yes \\ \hline
SMAC~\cite{smac}  & No & No & No  & 50       & No  & N/A & Limited (only new maps) \\ \hline
SMART~\cite{SMART} & Yes  & Yes & No  & 4        & Yes & No  & Yes                  \\ 
 &   &  &   &         &  &   &                   \\ \hline
MultiRoboLearn\cite{multirobolearn}  & Yes & Yes & Yes  & 4  & Yes  & No  & Yes, but Difficult         \\ \hline
Robotarium~{\cite{robotarium}}      & Yes & Yes (CBFs) & No   & 20 & Yes  & Yes & Yes, but Difficult                       \\ \hline
\textbf{MARBLER (ours)} & \textbf{Yes} & \textbf{Yes (CBFs)} & \textbf{Yes} & \textbf{20} & \textbf{Yes} & \textbf{Yes} & \textbf{Yes}  \\ \hline
\end{tabular}
\caption{Comparison of MARBLER with other platforms. MARBLER is the only MRRL platform with Sim2Real capabilities that allows for more than four robots and has a publicly available testbed.}
\label{table:comparison}
\end{table*}

The Multi-Agent Particle Environment (MPE)~\cite{MPE} is a popular framework for evaluating MARL algorithms, consisting of cooperative and adversarial 2D tasks. 
In MPE, agents apply forces to particles which can interact with landmarks and other agents. This is a popular setup in MARL environments and has been extended by platforms such as VMAS~\cite{vmas}: a vectorized MARL platform that can use GPUs for much faster training. However, particle simulators have very different dynamics than real robots making them poor choices to evaluate MRRL algorithms. 

Another popular MARL environment is the StarCraft Multi-Agent Challenge (SMAC)~\cite{smac} which is considerably more complex, requiring agents to handle partial observability over long horizons.
However, the agent dynamics in SMAC is still considerably different from real world robots, again making it a poor choice to evaluate MRRL algorithms. 


There are few frameworks that are designed to benchmark MRRL algorithms and fewer still that are able to evaluate Sim2Real performance of algorithms. SMART~\cite{SMART} is one such environment. However, SMART is limited to scenarios involving autonomous driving, it only supports up to four robots, and neither their evaluation test bed nor their source code is publicly available.
Another MRRL environment that allows for Sim2Real testing is MultiRoboLearn~\cite{multirobolearn}: an open-source framework that provides an OpenAI Gym interface for easier integration. However it also only supports a maximum of 4 robots, and, like SMART, does not have a publicly available testbed. Additionally, creating new scenarios in MultiRoboLearn requires creating custom environments in Gazebo \cite{gazebo}, introducing significant overhead.

In contrast to existing environments, MARBLER's simulator closely mimics the constraints of physical robots \emph{and} allows researchers to evaluate Sim2Real capabilities in a standardized and reproducible way. Therefore, MARBLER is the first MRRL benchmark that has a realistic simulator with a physical testbed that \emph{anyone} can use.

\subsection{MARL Algorithms} 
A variety of MARL algorithms have been proposed that perform very well in simulated environments. PPO~\cite{ppo} is an effective actor-critic policy gradient method for single agent RL. MAPPO~\cite{mappo} is the multi-agent extension of PPO where a single centralized critic is conditioned on all agents' observations to learn a joint state value function and a separate actor for each agent learns the best action to take conditioned only on the agent's individual observations.

In contrast to MAPPO, QMIX~\cite{qmix} and VDN~\cite{VDN} are value-based methods that decompose the joint state-action value function into individual state-action value functions. VDN learns to decompose the team value function agent-wise while QMIX learns agent-specific Q networks and combines them monotonically via hypernetworks.

In SMAC and MPE, MAPPO, QMIX, and VDN have been shown to be three of the best performing MARL algorithms~\cite{papoudakis2021benchmarking}.
However, while these algorithms have performed very well in simulation, there is limited evaluation of their real world performance.~\cite{SMART} evaluated VDN's and QMIX's performance on robots and~\cite{hetgppo} and~\cite{yu2023asynchronous} evaluate different versions of multi-agent PPO based algorithms on real robots. However, these experiments only used at most four robots and are not easily reproducible.

Another important design problem in MRRL is whether robots should share parameters. When robots share parameters, their networks all learn together which greatly reduces the number of parameters required to be trained. However, this leads to all robots learning the same behavior, rendering parameter sharing unsuitable in scenarios that require heterogeneous behaviors from robots.  
A common solution to combat this involves appending unique IDs to robots' observations, but the level of heterogeneity enabled by this approach has been shown to be limited~\cite{hetgppo}. Alternatively, each robot can learn its own set of network parameters which allows robots to learn truly heterogeneous behavior but limits scalability as it greatly increases the number of environment interactions needed for the robots to learn.

\subsection{The Robotarium}
The Robotarium\cite{robotarium} is a remotely accessible multi-robot laboratory developed by Georgia Tech. It features a 12ft x 14ft testbed, 8 Vicon motion-capture cameras and allows up to 20 GRITSBots~\cite{grits} to operate at once.  
The Robotarium has optional inbuilt control barrier functions (CBFs)~\cite{barrier} which provide a provable guarantee of online collision avoidance for the robots.
The Robotarium also provides a Python simulator that closely resembles the real Robotarium. Once programs are working in simulation, the Robotarium has a publicly accessible website where anyone in the world can upload their programs for them to be run in the real Robotarium on real robots.

\section{The MARBLER Platform}
Historically, evaluating MRRL algorithms using the Robotarium's simulator has been a challenging task. The lack of a standardized framework for MRRL in the Robotarium means that researchers have to create scenarios from scratch, design the low level control algorithms to control the robots after they select an action, control how the graphics are displayed, and more. As a result, to the best of our knowledge, only~\cite{haksar2018distributed} has evaluated deep reinforcement learning algorithms with the Robotarium, despite its open accessibility to researchers. Addressing this limitation, MARBLER establishes a cohesive and user-friendly API tailored specifically for MRRL experiments. Researchers can design novel scenarios or employ the pre-existing scenarios to execute their algorithms, thereby allowing reproducibility across studies.

After training and evaluating a policy in simulation, MARBLER enables researchers to generate the required files for deployment on the physical Robotarium with a single script. 

\begin{figure*}[t]
  \centering

 \begin{subfigure}{0.3\textwidth}
    \centering
    \includegraphics[width=\textwidth]{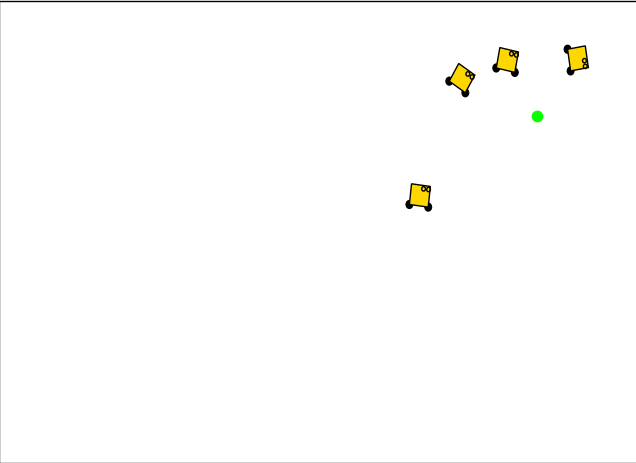}
    \includegraphics[width=\textwidth]{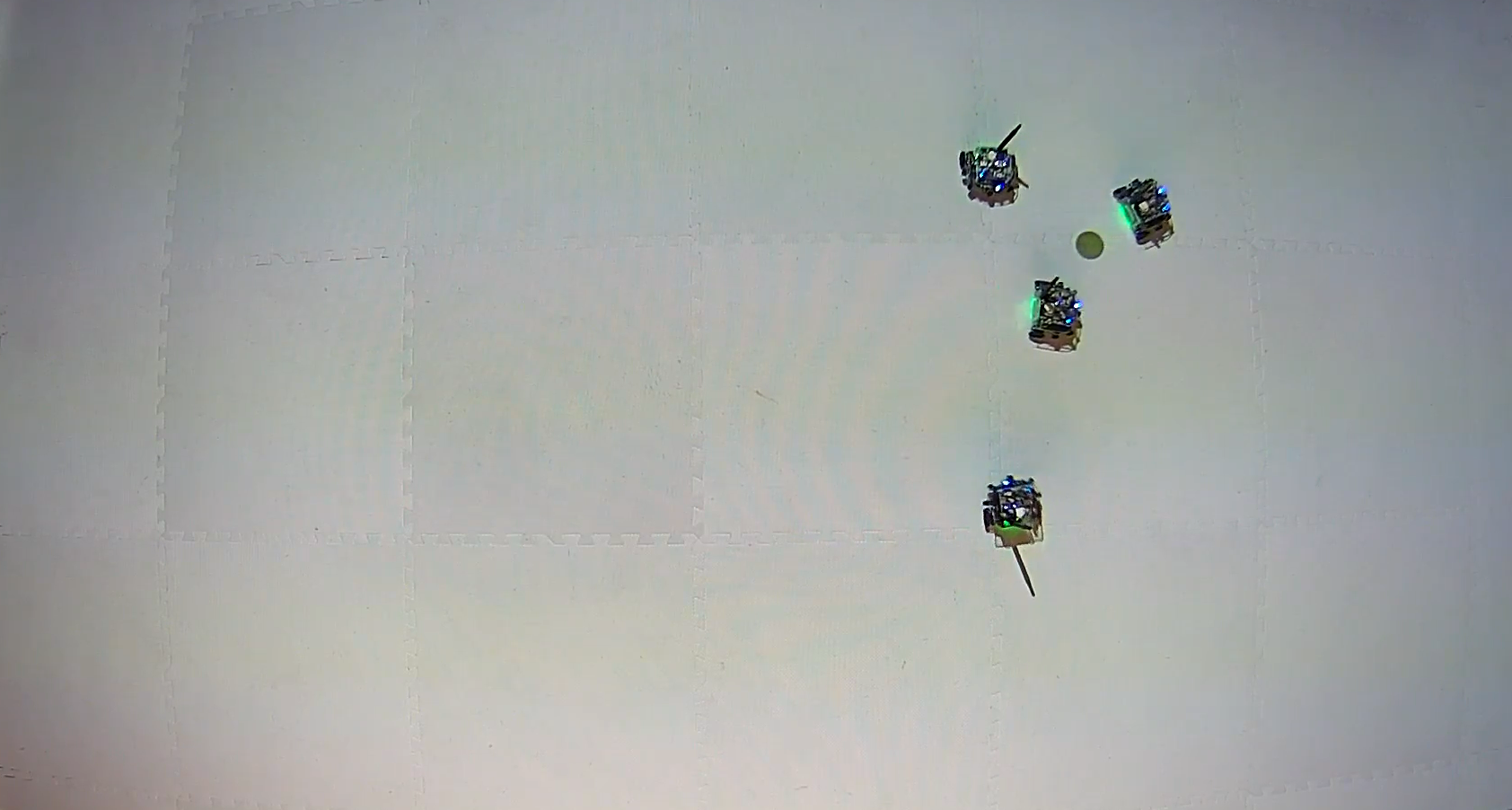}
    \caption{Simple Navigation}
    \label{fig:Simple}
  \end{subfigure}
  \begin{subfigure}{0.3\textwidth}
    \centering
    \includegraphics[width=\textwidth]{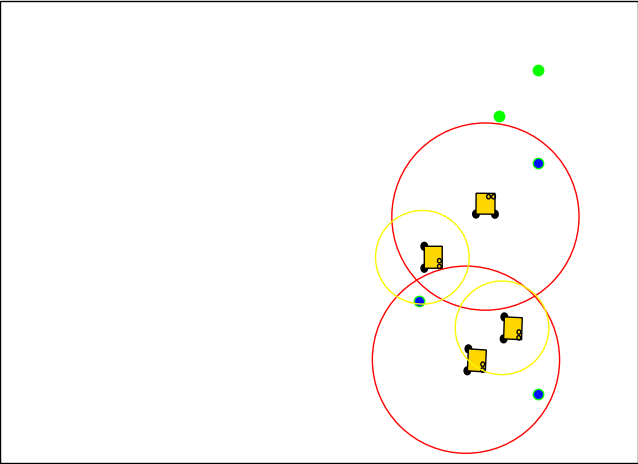}
    \includegraphics[width=\textwidth]{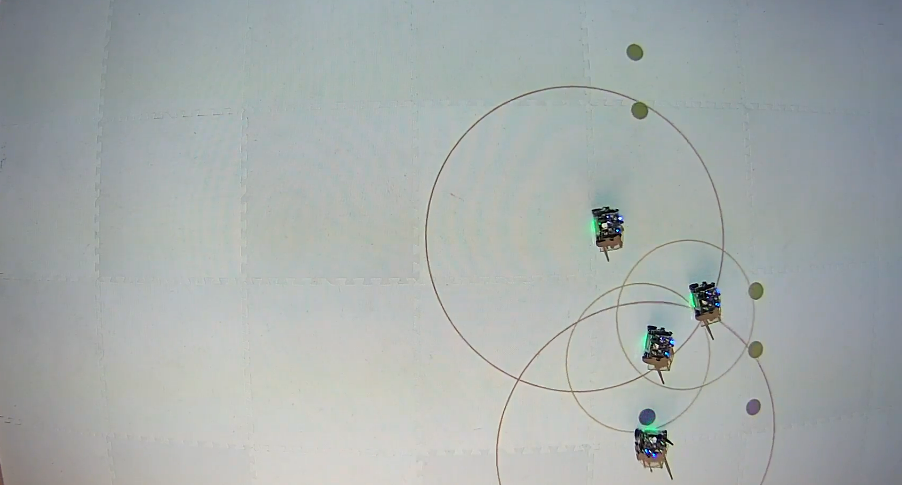}
    \caption{Predator Capture Prey}
    \label{fig:PCP}
  \end{subfigure}
  \begin{subfigure}{0.3\textwidth}
    \centering
    \includegraphics[width=\textwidth]{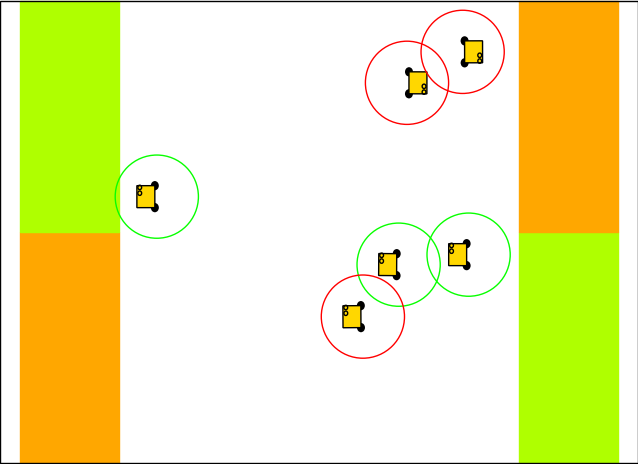}
    \includegraphics[width=\textwidth]{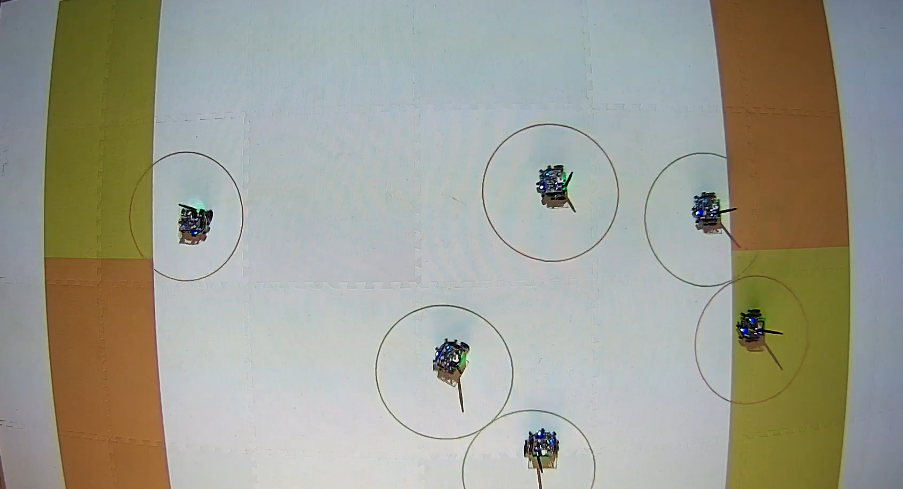}
    \caption{Warehouse}
    \label{fig:Warehouse}
  \end{subfigure}

  \vspace{.05cm}

  \begin{subfigure}{0.3\textwidth}
    \centering
    \includegraphics[width=\textwidth]{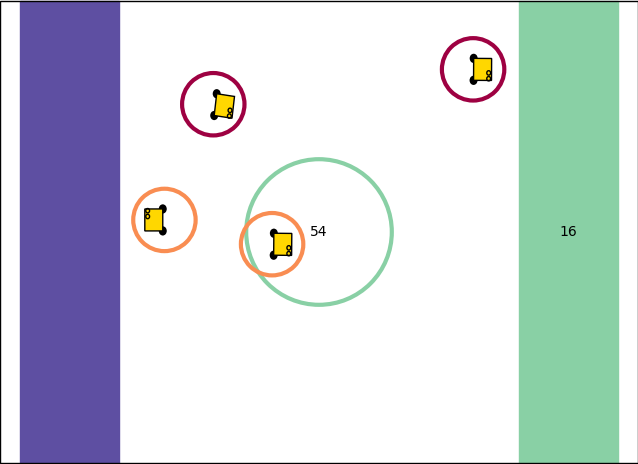}
    \includegraphics[width=\textwidth]{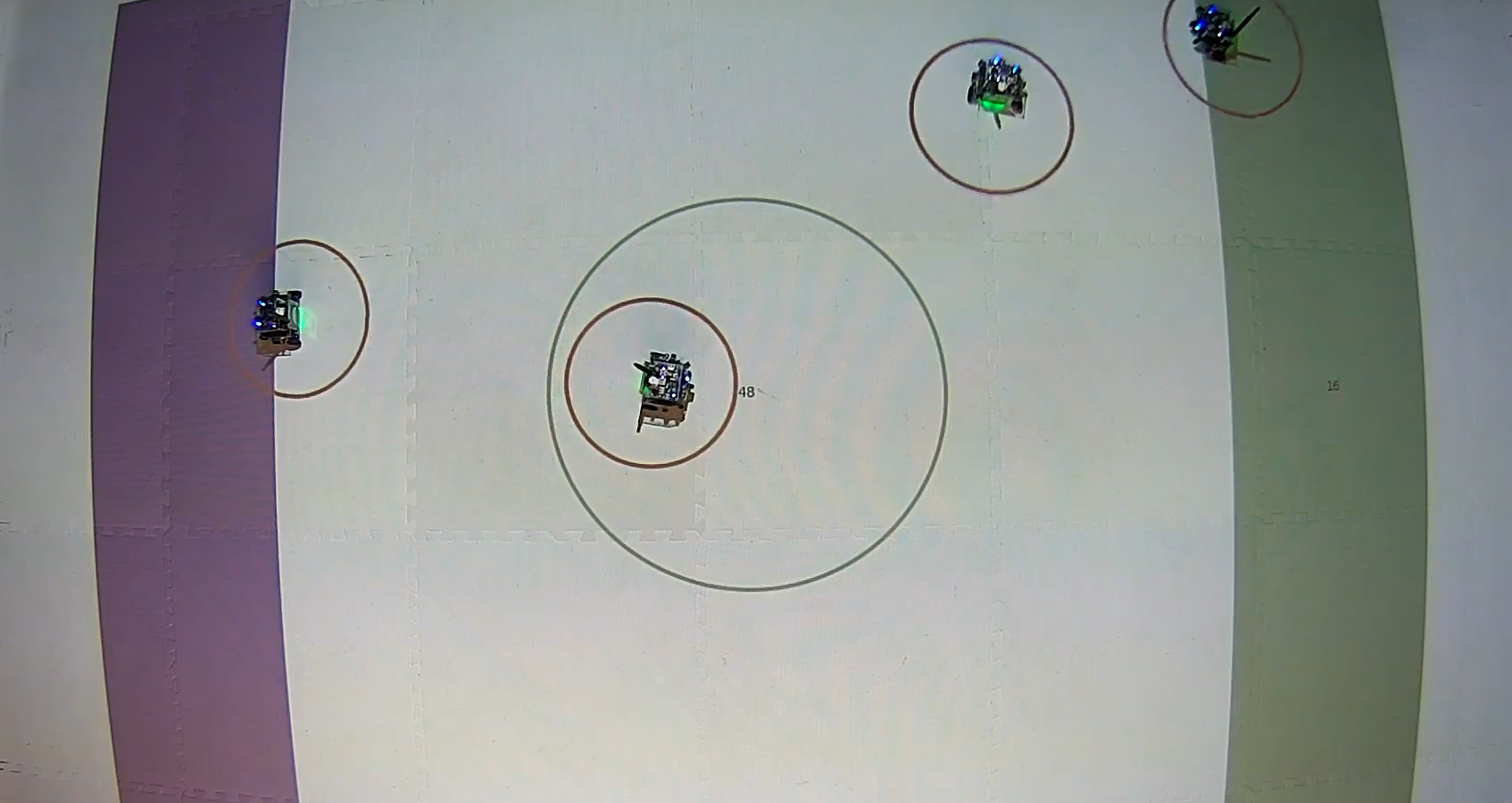}
    \caption{Material Transport}
    \label{fig:MaterialTransport}
  \end{subfigure}
  \begin{subfigure}{0.3\textwidth}
    \centering
    \includegraphics[width=\textwidth]{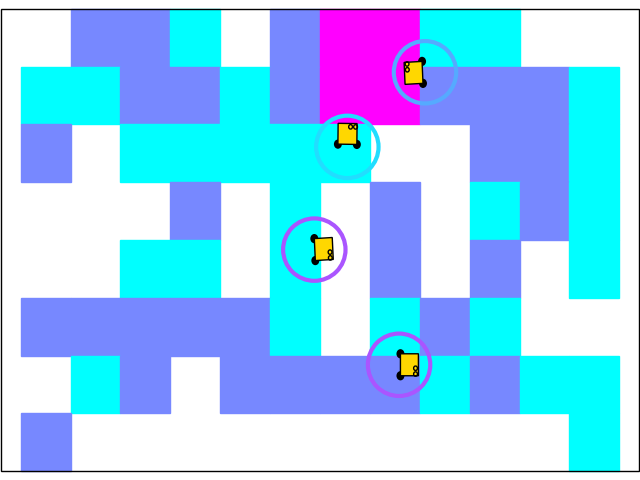}
    \includegraphics[width=\textwidth]{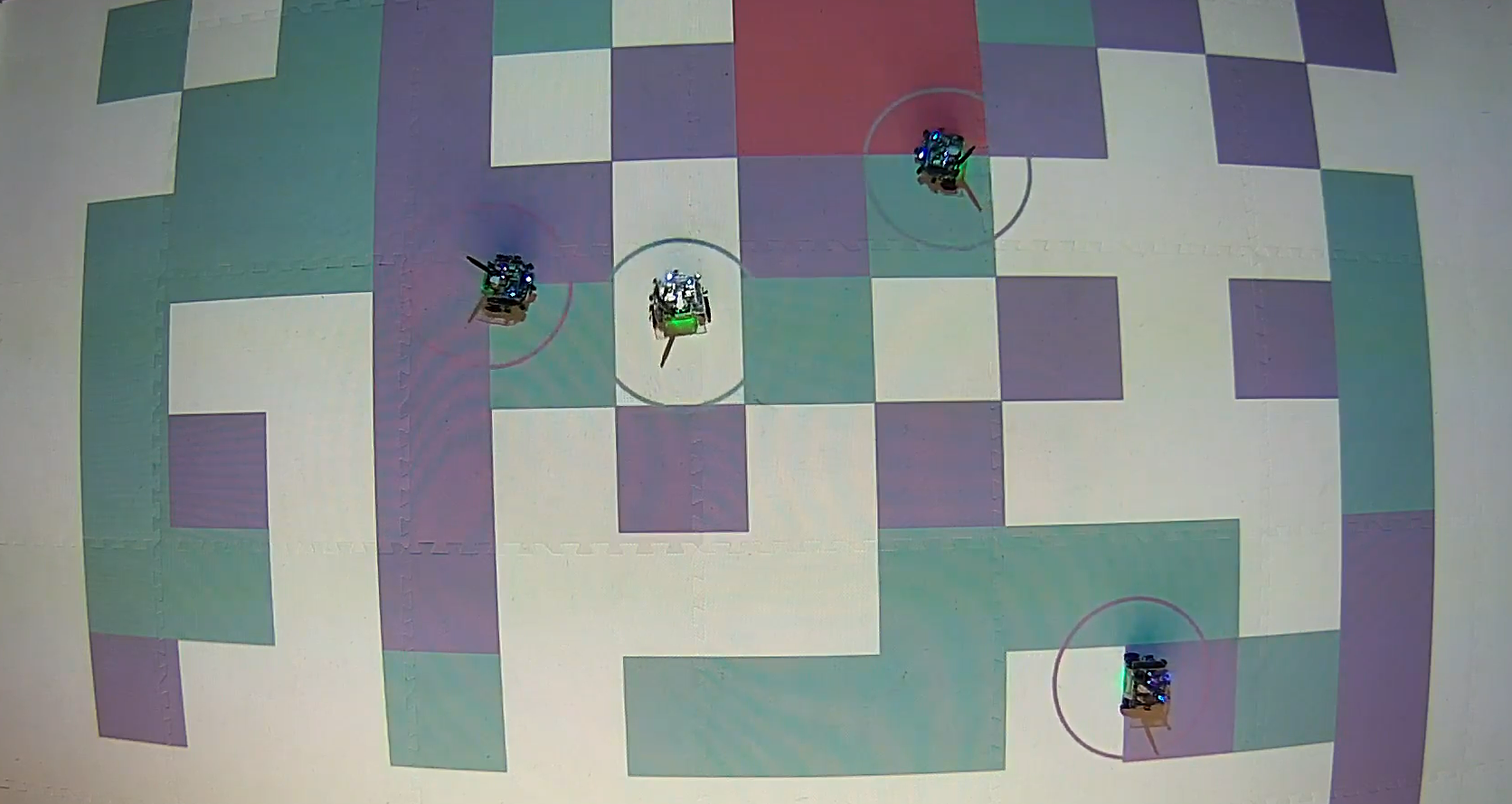}
    \caption{Arctic Transport}
    \label{fig:ArcticTransport}
  \end{subfigure}

  \caption{The existing scenarios in MARBLER. The top images show the robots running in simulation and the bottom images show the robots running in the Robotarium.}
  \label{fig:combined}
\end{figure*}

\subsection{Primary Components}
MARBLER is comprised of four primary components: 

\textbf{Core:} The Core component serves as the fundamental building block of MARBLER, leveraging the Robotarium's python simulator. It encompasses critical functionalities necessary for the environment, such as environment resets, CBF setup, and discrete time step advancement. By utilizing the capabilities of the Robotarium's simulator and CBFs, MARBLER incorporates realistic dynamics that emulate the constraints encountered by real robots.

\textbf{Scenarios:} The scenarios module defines the environments the robots interact in, the specific tasks they must accomplish, and the robots' observations and rewards.

\textbf{Gym Interface:}  Each scenario in MARBLER is registered as a Gym environment, enabling direct compatibility with any algorithms and tools that support the Gym interface.

\textbf{Test Pipeline:} The Test Pipeline
provides a streamlined process for importing trained robots into the simulation environment, giving researchers a way to visualize robots' performance and collect test data. Subsequently, researchers can execute a script to prepare their files for submission to the Robotarium, which can then be uploaded to the real Robotarium for evaluation in a real-world setting.

\subsection{Scenarios}
\subsubsection{Existing Scenarios}

To facilitate immediate testing and evaluation using MARBLER, we introduce five scenarios inspired by diverse MRRL problems. These scenarios are designed to offer researchers a starting point for experimentation and can be easily customized by modifying the scenario's associated configuration file. Parameters such as the number of robots, communication methods, scenario difficulty, and more, can be adjusted as needed.
Complete descriptions of these scenarios are available in the supplementary material\footnote{Supplementary material can be found
\href{https://shubhlohiya.github.io/MARBLER/assets/supplementary.pdf}{here}}
but we include brief descriptions here: 

\textbf{Simple Navigation (Fig. \ref{fig:Simple}):} 
Robots navigate towards a known destination point. This scenario is an easy starting point for algorithms to learn in.

\textbf{Predator Capture Prey (PCP) (Fig. \ref{fig:PCP}):} 
Sensing robots and capture robots must work together to capture the prey. Sensing robots know the location of prey within their sensing radius and must communicate this to the blind capture robots. Inspired by the Predator Capture Prey scenario in~\cite{hetnet}.

\textbf{Warehouse (Fig. \ref{fig:Warehouse}):} 
Robots must navigate to their color zone on the right to receive a load and then unload in their color zone on the left while avoiding collisions; a Multi-Robot Path Finding environment~\cite{mapfDef}. 

\textbf{Material Transport (MT) (Fig. \ref{fig:MaterialTransport}):}
Robots with varying speeds and capacities must collaborate to efficiently unload two zones: one nearby with a large amount of material and one further away with a small amount of material. This is a task allocation problem~\cite{neville2021interleaved} where the robots must collaborate to unload the zones within a time limit.

\textbf{Arctic Transport (AT) (Fig. \ref{fig:ArcticTransport}):}
Drones can move fast over any tile and have a large sensing radius. Ice and water robots have a limited sensing radius and move fast over some tiles but slow over other tiles. Robots are rewarded based on how far the ice/water robots are from the goal zone so the drones must guide the ice/water robots. This is a Multi-Robot Path Planning scenario~\cite{mapf} where the drones must find a path to the goal zone and communicate it to the ice/water robots.

\subsubsection{Creating New Scenarios}
MARBLER provides a user-friendly approach to create new scenarios, similar to MPE and VMAS. Researchers can customize the action space, observation space, visualizations, and other relevant parameters without needing to interact with the underlying Robotarium code, allowing researchers to develop tailored scenarios that align with their specific use cases. Our GitHub\footnote{
Our GitHub can be found \href{https://github.com/GT-STAR-Lab/MARBLER}{here}
} includes comprehensive documentation on creating new scenarios.

\section{Experiments}\label{Experiments}
\subsection{Experiment Setup}
\begin{figure*}[h]
  \centering
  \includegraphics[width=\textwidth]{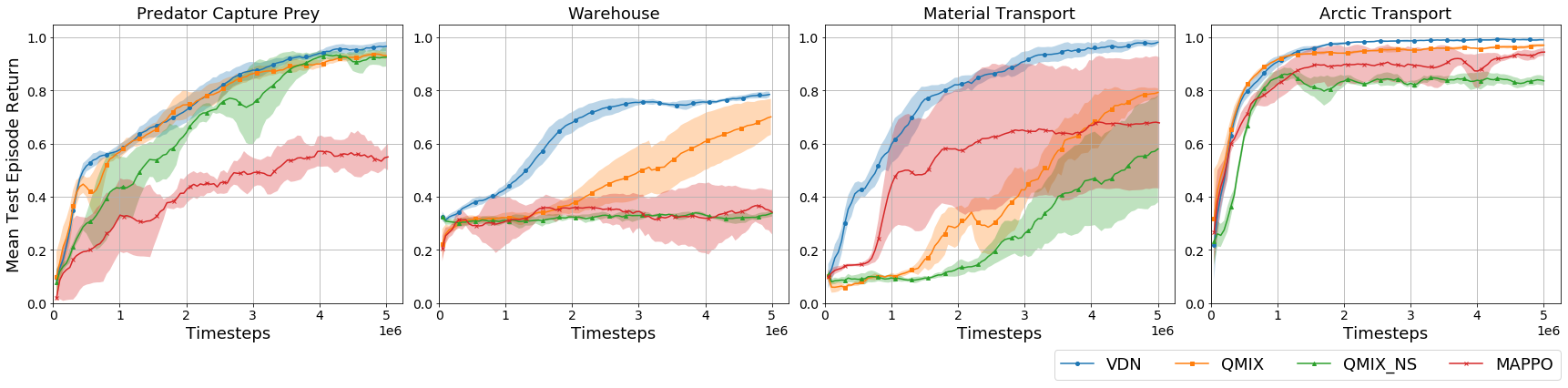}
  \caption{Evaluation returns for each algorithm during training. The lines are the mean rewards across the five seeds and the shaded area is the 95\% confidence interval. The remaining time steps for MAPPO can be seen in the supplementary material.}
  \label{fig:learning_curves}
\end{figure*}

For all our experiments, we used the EPyMARL framework to train our robots. Because the scenarios in MARBLER have been registered as Gym environments, they are directly compatible with EPyMARL. This allowed us to train policies using the various learning algorithms available in EPyMARL with no modifications.

\textbf{Baselines}: We compared MAPPO~\cite{mappo}, QMIX~\cite{qmix}, and VDN~\cite{VDN} with parameter sharing. To investigate the effects of parameter sharing, we also evaluated QMIX without parameter sharing (QMIX\_NS).

\subsection{Evaluation Protocol}
We evaluated all algorithms in the PCP, Warehouse, MT, and AT scenarios with 4, 6, 4, and 4 robots respectively. Before training each algorithm, we ran a hyperparameter search in the Simple Navigation environment in a manner similar to~\cite{papoudakis2021benchmarking}. Exact details on the hyperparameter search along with the hyperparameters we used for each algorithm can be found in the supplementary material.

We trained VDN and QMIX for a total of 5 million time steps in each scenario. Given the conflicting evidence about off-policy algorithms being more sample efficient than on-policy algorithms due to their use of a replay buffer~\cite{mappo,papoudakis2021benchmarking}, we trained MAPPO for a total of 25 million time steps. We trained five seeds for each algorithm.

Researchers control what happens if robots go outside of the Robotarium's boundaries or collide in simulation. However, the real Robotarium immediately terminates an episode if either event happens. To prevent frequent and premature termination in the Robotarium, we used strict CBFs such that, if the robots attempt to get within 20cm from each other, their movement slows down to almost a full stop. Like the real Robotarium, we penalize the robots and end the episode if robots collide or drive out of bounds.

In all scenarios, robots had full communication. Exact details about how the environments were configured for these evaluations are included in the supplementary material.

\subsection{Computational Requirements}
We trained all models using CPUs; primarily with a Dual Intel(R) Xeon(R) Gold 6226~\cite{PACE} and an Intel(R) Core(TM) i7-12700KF. It took 16084 CPU hours to train all models (excluding hyperparameter searches).

\section{Results}

\begin{table*}[t]
\centering

\resizebox{\textwidth}{!}{%
\begin{tabular}{cl|cccc|cccc}
        \toprule
            & & \multicolumn{4}{c|}{\textbf{Simulated Experiments}} & \multicolumn{4}{c}{\textbf{Real-World Experiments}} \\ \cline{3-10} & & & & & & & & & \\[-1.9ex]
        \textbf{Scenario} & \textbf{Metric} & \textbf{MAPPO} & \textbf{VDN} & \textbf{QMIX} & \textbf{QMIX\_NS} & \textbf{MAPPO} & \textbf{VDN} & \textbf{QMIX} & \textbf{QMIX\_NS} \\
        \hline & & & & & & & & & \\[-1.9ex]
         \multirow{4}{*}{\parbox{1.4cm}{\centering \textbf{Predator Capture Prey}}}
        & Reward        & 23.48$\pm$5.33   & \textbf{33.25$\pm$0.46} & 30.02$\pm$3.8  & 31.76$\pm$2.7  & 21.63$\pm$9.06 & \textbf{31.51$\pm$5.51} & 29.2$\pm$4.4 & 30.13$\pm$5.26  \\
        & Steps         & 80.4$\pm$1.8     & \textbf{55$\pm$9.18}    & 69.7$\pm$12.32 & 62.9$\pm$12.75 & 81$\pm$0       & \textbf{57.8$\pm$15.79} & 70$\pm$13.63 & 67.5$\pm$11.38 \\
        & Prey Left     & 1.6$\pm$0.92     & \textbf{0$\pm$0}        & 0.5$\pm$0.67   & 0.2$\pm$0.4    & 2$\pm$1.63     & \textbf{0.3$\pm$0.95}   & 0.6$\pm$0.7  & 0.5$\pm$0.97 \\
        & Collisions    & 0                & 0                       & 0              & 0              & 10\%           & 0                       & 0            & 0 \\ 
        & Out of Bounds & 0                & 0                       & 0              & 0              & 0              & 0                       & 0            & 0 \\ 
        \hline & & & & & & & & & \\[-1.9ex]
         \multirow{2}{*}{\parbox{1.4cm}{\centering \textbf{Warehouse}}}
        & Reward        & \textbf{36.6$\pm$1.8} & 28.7$\pm$1.49 & 27.4$\pm$1.02 & 1.8$\pm$1.25 & \textbf{35.1$\pm$2.47} & 26.2$\pm$0.79 & 26.89$\pm$1.76 & -3.21$\pm$11.18 \\
        & Collisions    & 0                     & 0             & 0             & 0            & 0                      & 0             & 0              & 0 \\ 
        & Out of Bounds & 0                     & 0             & 0             & 5\%          & 0                      & 0             & 0              & 20\% \\ 
        \hline & & & & & & & & & \\[-1.9ex]
         \multirow{4}{*}{\parbox{1.4cm}{\centering \textbf{Material Transport}}}
        & Reward        & 4.47$\pm$0.93  & \textbf{5.15$\pm$1.3}  & 3.55$\pm$0.85 & 2.08$\pm$0.85   & 3.76$\pm$2.19 & \textbf{5.73$\pm$1.16} & 3.72$\pm$1.14 & 1.78$\pm$1.97 \\
        & Steps         & 71$\pm$0      & \textbf{65.5$\pm$4.48} & 71$\pm$0      & 71$\pm$0        & 70.4$\pm$1.26 & \textbf{60.1$\pm$7.09} & 71$\pm$0      & 71$\pm$0 \\
        & Material Left & 8.4$\pm$4.41  & \textbf{0.1$\pm$0.3}     & 9.4$\pm$4.59  & 28.70$\pm$12.74 & 8.7$\pm$13.51 & \textbf{0.1$\pm$0.32}   & 14$\pm$10.14  & 32.6$\pm$12.36 \\
        & Collisions    & 0             & 0                      & 0             & 0               & 10\%          & 0                      & 0             & 0 \\ 
        & Out of Bounds & 1\%           & 0                      & 0             & 4\%             & 0             & 0                      & 0             & 10\% \\ 
        \hline & & & & & & & & & \\[-1.9ex]
         \multirow{3}{*}{\parbox{1.4cm}{\centering \textbf{Arctic Transport}}}
        & Reward        & -7.23$\pm$1.61 & -6.98$\pm$1.75 & -7.13$\pm$1.59          & -11.29$\pm$3.29 & -7.91$\pm$1.66 & -7.86$\pm$3.33          & -12.15$\pm$9.8 & -18.49$\pm$13.34 \\
        & Steps         & 41.7$\pm$10.65 & 38.1$\pm$10.25 & \textbf{35.8$\pm$8.24}  & 57$\pm$7.71     & 46.5$\pm$15.92 & \textbf{34.4$\pm$11.15} & 43.6$\pm$13.07 & 51.4$\pm$13.04 \\
        & Collisions    & 0              & 0              & 0                       & 0               & 0              & 0                       & 10\%           & 0 \\ 
        & Out of Bounds & 0              & 0              & 0                       & 1\%             & 10\%           & 0                       & 0              & 30\% \\ 
               
        \bottomrule
    \end{tabular}
    %
    }
    \caption{The mean returns and standard deviations of each algorithm for every scenario. The simulated results were taken over 100 episodes and the results from real robots were taken across 10 episodes. Collisions refer to the percent of episodes terminated due to robots colliding, Out of Bounds refers to the percent of episodes terminated due to robots going outside the boundary of the Robotarium. The steps for episodes that end due to a collision or a boundary violation is set to the maximum. Best values for simulation and real in each row are bolded. Note that robots never collide in simulations.}
    \label{table:evals}
\end{table*}

To compare baselines, first we look at training evaluation returns to evaluate sample efficiency and how much of an impact different seeds make which can be seen in Fig. \ref{fig:learning_curves}. Then, we compared the best performing models for each algorithm in each scenario. To do this, we took the model that achieved the highest reward for each algorithm and evaluated the model in simulation and on real robots to compare performances. In simulation, we ran each model for 100 episodes and on the real robots, we ran each model for 10 episodes. The results can be seen in table \ref{table:evals}.

\subsection{Value Based vs. Policy Gradient}
VDN outperforms all other algorithms after 5 million time steps for every scenario. After 25 million steps, MAPPO's best performing seeds approaches that of VDN's in MT and AT and surpasses it in Warehouse. However, all seeds for MAPPO converge to lower performance in PCP than in any of the value based methods.
Additionally, MAPPO's performance is much more sensitive to the random seed than any value-based method. This is contradictory to the trends reported in \cite{mappo}.
We speculate this is because value based methods (particularly VDN) may be more suitable to physical robots than policy gradient methods. 

\subsection{Effects of Parameter Sharing}
We find that the differences between models trained with and without parameter sharing depend on the heterogeneity of the environment. In the Warehouse scenario, where robots are homogeneous except for their loading zone locations, QMIX outperformed QMIX\_NS significantly. In MT, the robots need to learn slightly different policies to ensure that all zones are unloaded within the time limit, but the optimal policies are similar. In AT, drones and ice/water robots had fundamentally different optimal policies, yet neither QMIX nor QMIX\_NS utilized the drones' enhanced sensing radius, resulting in similar policies for all robots. In AT and MT, with limited heterogeneity, QMIX showed a significant performance advantage over QMIX\_NS but much less significant than in Warehouse. However, in the PCP scenario, where very different policies were learned for the Predator and the Capture robots, QMIX and QMIX\_NS performed similarly. Thus, the benefits of parameter sharing evaporate as heterogeneity increases. Indeed, as previously reported~\cite{hetgppo}, models trained without parameter sharing tend to outperform models that share parameters in increasingly heterogeneous environments. 

Additionally, robots trained with QMIX\_NS went out of bounds a total of 10 times in simulation and 6 times on real robots. In contrast, robots trained with \emph{all} parameter sharing methods only went out of bounds once in simulation and once on real robots. When a single robot goes out of bounds, all robots are given a large negative penalty and the episode ends. 
This suggests that, without parameter sharing, it is much more difficult for robots to learn how to handle events where a single robot's actions can cause all other robots to suffer a penalty. 

\subsection{Sim2Real Gap}

As shown in table \ref{table:evals}, there are few notable differences between the algorithms' average
performances in simulation and in the real Robotarium. Additionally, the difference in standard deviations can be partly attributed to the fact that we have 100 simulated episodes, but only 10 real episodes. These observations suggest the lack of a significant Sim2Real gap. However, there is one key difference between the real and simulated experiments: the robots never collide in simulation and the robots go out of bounds more than 6x more often on real robots. The only time an algorithms' metrics were significantly worse on real robots vs. in simulation was when the real robots collided or went out of bounds.

To further evaluate the Sim2Real gap, we compared our VDN policies in PCP against a new version of VDN with two crucial modifications that mimic the default safety mechanisms of the Robotarium. First, we used CBFs that are only effective at 17cm and do not slow the robots as much when they are within the safety radii. Second, we did not terminate the episode or penalize the robots for driving out of bounds or colliding. We refer to this as VDN (Default CBF). To ensure a fair comparison, we penalized the robots for colliding or driving out of bounds during testing, even though VDN (Default CBFs) was not penalized during training.

\begin{table}[h]
\centering
\begin{tabular}{cl|cc}
        \toprule
        \textbf{Scenario} & \textbf{Metric} & \textbf{VDN (Safe CBF)} & \textbf{VDN (Default CBF)} \\

        \hline & & & \\[-1.7ex]
         \multirow{4}{*}{\parbox{1.2cm}{\centering \textbf{Predator Capture Prey}}}
        & Reward & 33.25$\pm$0.46 & 30.34$\pm$4.63  \\
        & Steps & 55$\pm$9.18 & 63.20$\pm$12.39 \\
        & Prey Left & 0$\pm$0 & 0.10$\pm$0.30 \\
        & Collisions & 0 & 0 \\ 
        & Boundaries & 0 & .03 \\         
        \bottomrule
    \end{tabular}
    \caption{Comparison of VDN (Safe CBF) against VDN (Default CBFs) in simulation.}
    \label{table:ablation}
\end{table}

As seen in table \ref{table:ablation}, there are no significant differences in simulation between the performance of VDN (Safe CBFs) and VDN (Default CBFs). When evaluated on the physical Robotarium, no episode of VDN (Safe CBFs) resulted in a collision.
However, VDN (Default CBFs) resulted in collisions 3 out of 10 episodes despite using the recommended safety mechanisms. Note that 100 simulated episodes of VDN (Default CBFs) had no collisions. These findings suggest a significant Sim2Real gap when it comes to safety: even if robots seem to learn safe policies in simulation, we cannot assume safety in the real world. This observation highlights another significant contribution of MARBLER: it is the first open platform that can evaluate the safety of learned MRRL policies directly in the real world.

\section{Limitations and Future Work}
MARBLER's primary limitation is its training speed.
On an Intel i7-12700H, MPE trains $\sim$3.9 times faster than MARBLER with CBFs and $\sim$2.8 times faster than MARBLER without CBFs.
Future work will focus on improving MARBLER's training speed. We also did not benchmark MARBLER with GNN-based communication algorithms to maintain a fair comparison with EPyMARL, which lacks GNN support in its implementations. However, MARBLER has already been employed in studies that use GNNs~\cite{cagnn}. Finally, MARBLER hasn't been evaluated with complex observations (e.g., images) so future research should create and analyze scenarios with more realistic observations.

\section{Conclusion}
We introduced MARBLER, the first open platform with Sim2Real capabilities, realistic robot dynamics, and the ability to evaluate the safety of MRRL algorithms in the real world. MARBLER environments are fully compatible with the OpenAI Gym interface, providing an easy interface for modern learning algorithms. 
We also created five MRRL scenarios in MARBLER and utilized the EPyMARL framework to benchmark popular MARL algorithms, both in simulation and in the real-world. We hope MARBLER will help researchers benchmark Sim2Real transfer capabilities of MRRL algorithms in a systematic and reproducible way.

\bibliographystyle{IEEEtran}
\bibliography{clean}

\end{document}